\documentclass[conference]{IEEEtran}
\IEEEoverridecommandlockouts
\usepackage{cite}
\usepackage{amsmath,amssymb,amsfonts}
\usepackage{algorithmic}
\usepackage{graphicx}
\usepackage{textcomp}
\usepackage{algorithm}
\usepackage{subfigure}
\usepackage{color}

\def\BibTeX{{\rm B\kern-.05em{\sc i\kern-.025em b}\kern-.08em
    T\kern-.1667em\lower.7ex\hbox{E}\kern-.125emX}}
\begin{document}
\title{An End-to-End Geometric Deficiency Elimination Algorithm for 3D Meshes\\
\thanks{The corresponding author is Prof. Yang Cong.
	 This work is supported by National Nature Science Foundation of China under Grant (61722311, U1613214, 61821005, 61533015) and Cooperation Projects of CAS \& ITRI (CAS-ITRI201905).}
}
\author{
	\IEEEauthorblockN{Bingtao Ma$^{1,2,3}$,
		 Hongsen Liu$^{1,2,3}$, Liangliang Nan$^{4}$, Xu Tang$^{1,2}$, Huijie Fan$^{1,2}$ and Yang Cong$^{1,2,*}$
	}
\IEEEauthorblockA{
	\textit{$^{1}$State Key Laboratory of Robotics, Shenyang Institute of Automation, Chinese Academy of Sciences, Shenyang, 110016, China.} \\
	\textit{$^{2}$Institutes for Robotics and Intelligent Manufacturing, Chinese Academy of Sciences, Shenyang, 110016, China.}\\
	\textit{$^{3}$University of Chinese Academy of Sciences,100049, China.}\\
	\textit{$^{4}$Delft University of Technology, Delft, 2628BL, Netherlands.}\\
mabingtao93@gmail.com, congyang81@gmail.com}}

\maketitle
\begin{abstract}
The 3D mesh is an important representation of geometric data. In the generation of mesh data,  geometric deficiencies (\emph{e.g.}, duplicate elements, degenerate faces, isolated vertices, self-intersection, and inner faces) are unavoidable and may violate the topology structure of an object.
In this paper, we propose an effective and efficient geometric deficiency elimination algorithm for 3D meshes. 
Specifically, \emph{duplicate elements} can be eliminated by assessing the occurrence times of vertices or faces; \emph{degenerate faces} can be removed according to the outer product of two edges; since \emph{isolated vertices} do not appear in any face vertices, they can be deleted directly; \emph{self-intersecting} faces are detected using an AABB tree and remeshed afterward;
by simulating whether multiple random rays  that shoot from a face can reach infinity, we can judge whether the surface is an \emph{inner face}, then decide to delete it or not.
Experiments on ModelNet40 dataset illustrate that our method can eliminate the deficiencies of the 3D mesh thoroughly.
\end{abstract}
\begin{IEEEkeywords}
3D Mesh; Geometric deficiencies; Mesh repair.
\end{IEEEkeywords}

\section{Introduction}

\begin{figure}
	\centering
	\subfigure[Duplicate elements.]
	{
		\label{fig:dup} 
		\includegraphics[width=0.45\columnwidth]{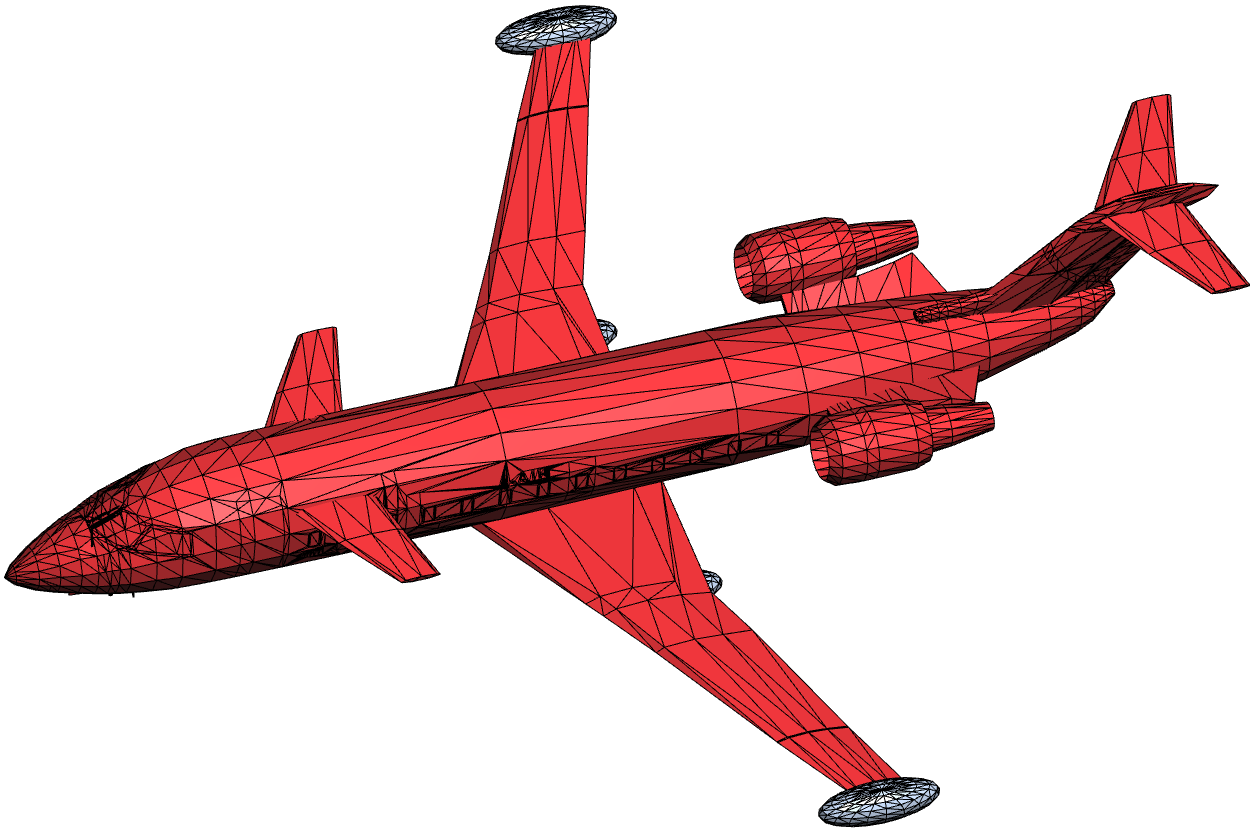}
	}
	\subfigure[Isolated vertices.]
	{
		\label{fig:iso} 
		\includegraphics[width=0.45\columnwidth]{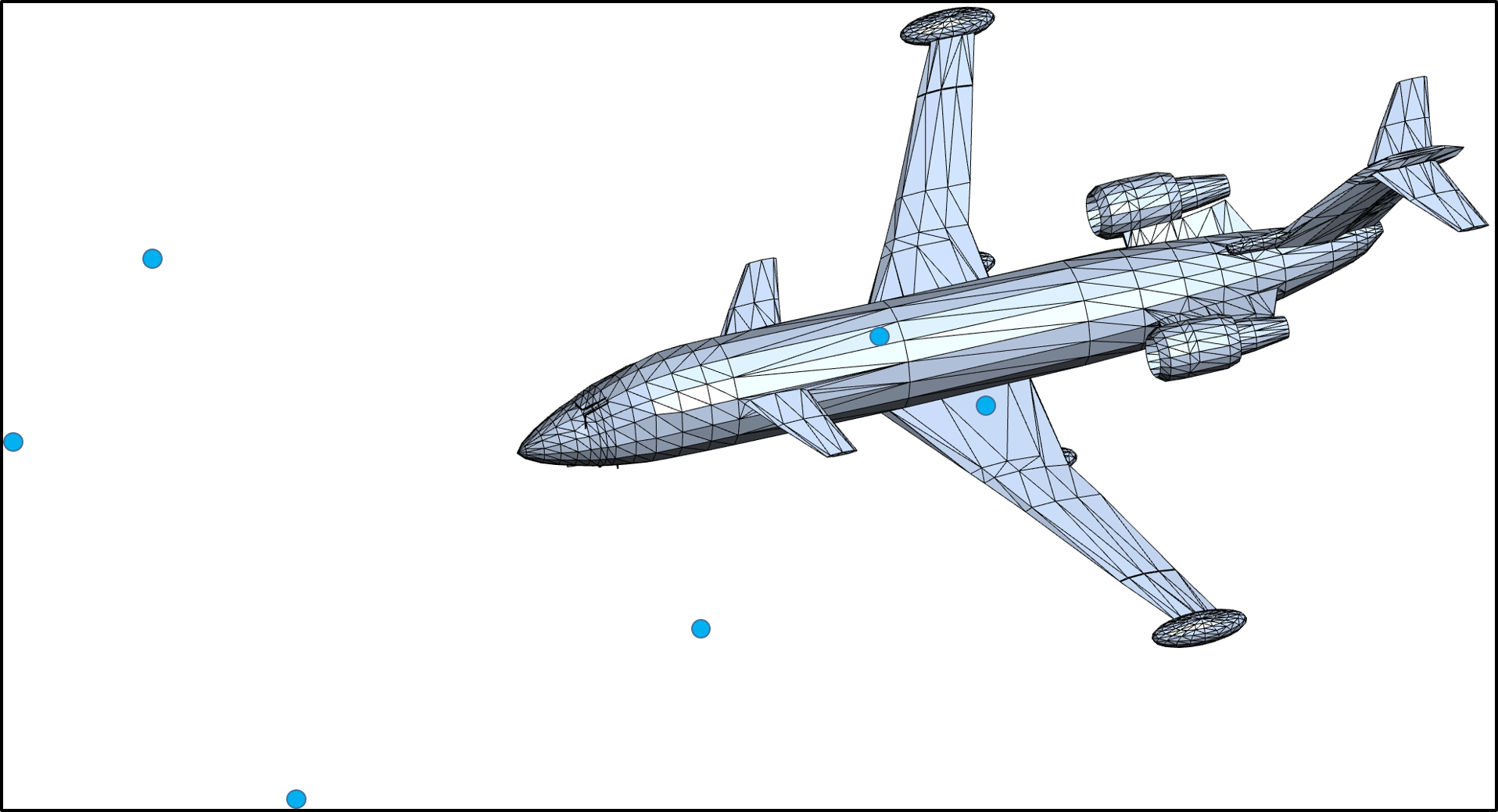}
	}
	\subfigure[Degenerate faces]
	{
		\label{fig:degenerate} 
		\includegraphics[width=3.4in]{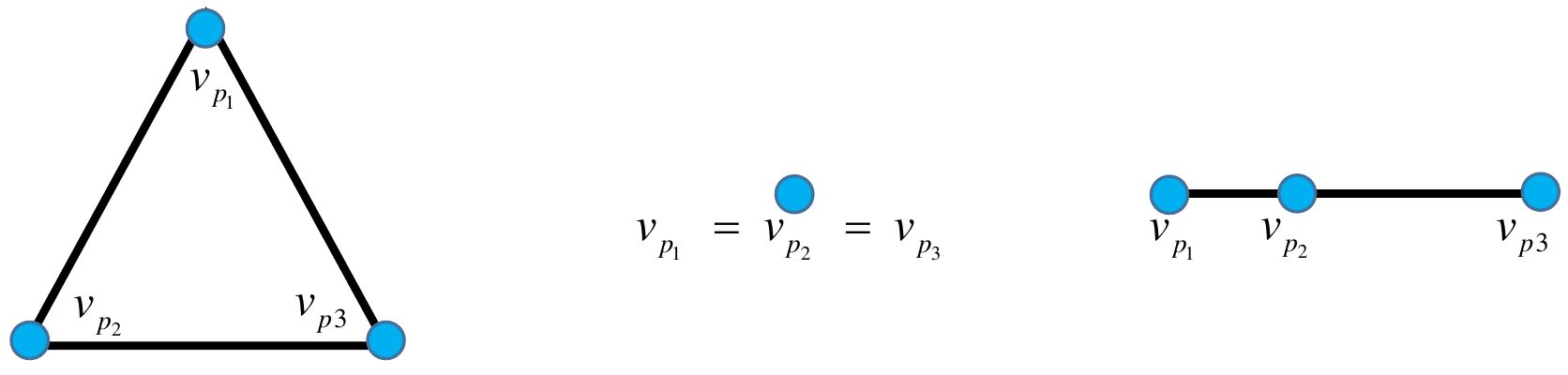}
	}
	\subfigure[Self-intersection.]
	{
		\label{fig:si} 
		\includegraphics[width=0.45\columnwidth]{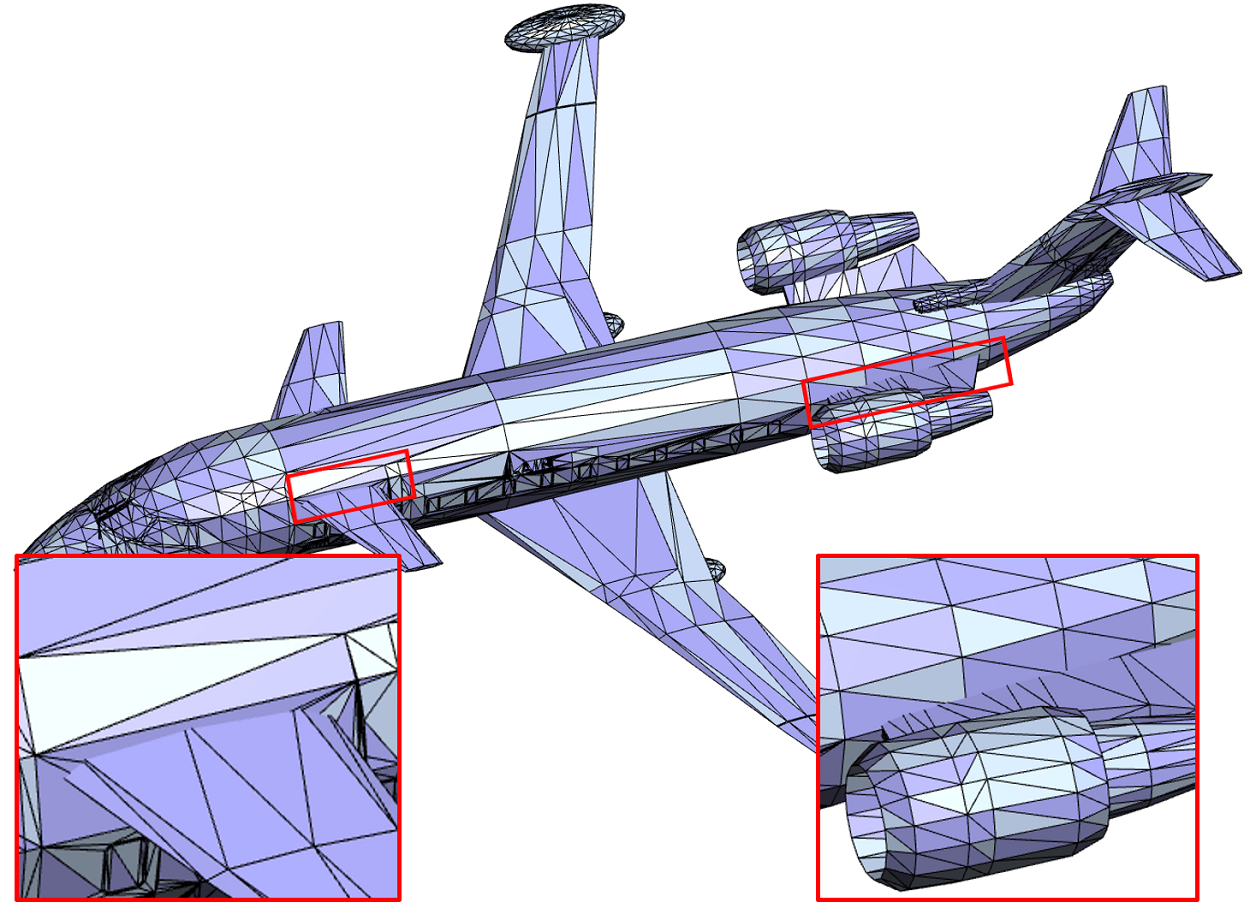}
	}
	\subfigure[Inner faces.]
	{
		\label{fig:inner} 
		\includegraphics[width=0.45\columnwidth]{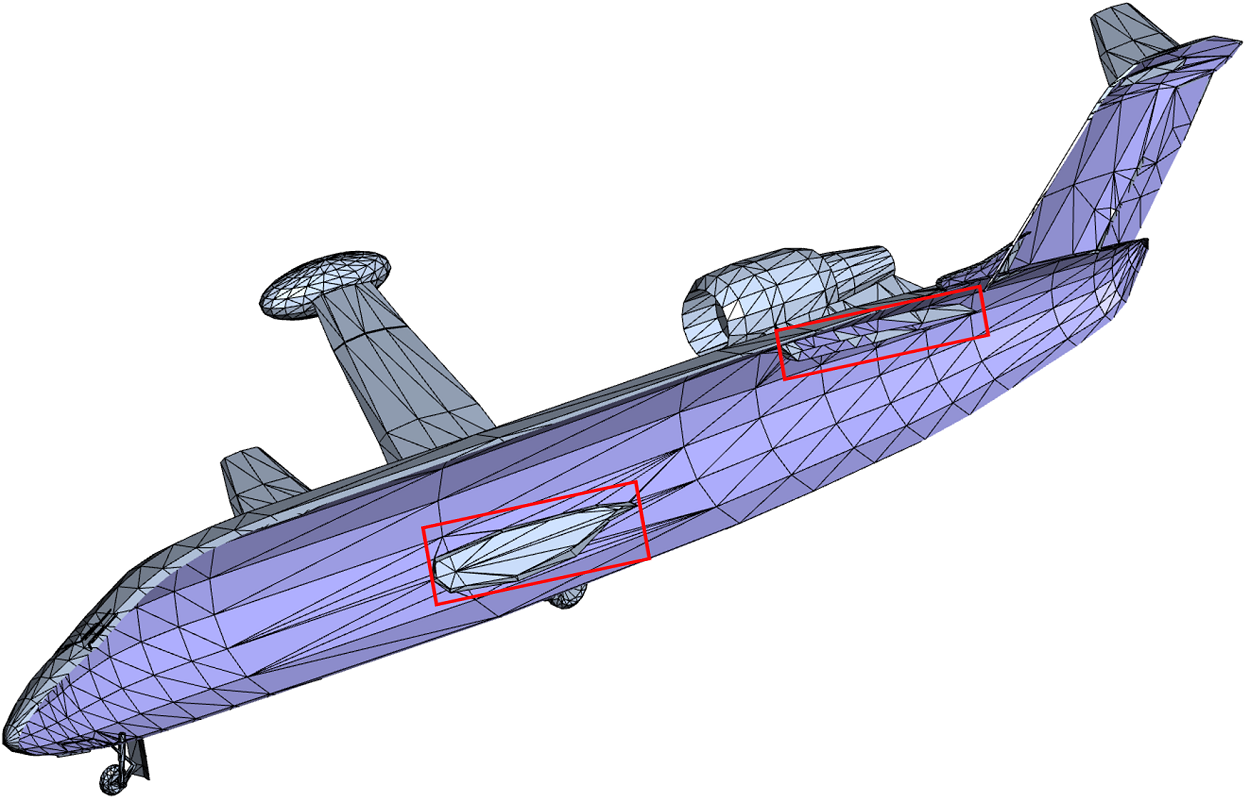}
	}
	\caption{ The common deficiencies of 3D meshes, which can be eliminated by the proposed method. (a) Duplicate elements are some vertices or faces that occur multiple times and are marked in red. (b) Isolated vertices are not used by any faces and are highlighted in blue. (c) Normal triangular face (left) and degenerate faces (middle and right). Some faces degenerate to a point (\emph{i.e.}, for a face $t_i$, $v_{p_{1}} = v_{p_{2}} = v_{p_{3}}$, where $v_{p_{1}}$, $v_{p_{2}}$, and $v_{p_{3}}$ are three vertices of a triangular face) or a line segment. (d) Self-intersection means their intersection is not an edge of the mesh. We visualize them in zoomed parts. (e) A cross-section view of the model. The marked faces are inner faces.}
	\vspace{-10pt}
	\label{fig:deficiencies} 
\end{figure}

Since 3D meshes have many advantages (\emph{e.g.}, rich topological information, lightweight, compact, and easier geometric transformation), 3D computer vision applications begin to pay more attention to mesh data, such as object reconstruction \cite{pixel2mesh} \cite{3drcnn}, scene reconstruction \cite{meshPiazza2018} \cite{Rosinol_meshre} and object recognition \cite{Deepshape} \cite{wavekernel} \cite{L3DOC}. However, most meshes have a lot of geometric deficiencies when they are generated, \emph{e.g.}, ModelNet40 \cite{modelnet} and ShapeNet \cite{shapenet}. The common geometric deficiencies are presented in Fig. \ref{fig:deficiencies}.
(a) Duplicate elements are some vertices or faces that appear repeatedly. Some structures of the mesh should have been topologically connected, but the duplicate elements cause them just overlaid together and cause topology connections erroneous.
(b) Isolated/Unreferenced vertices are not used by any faces, which may affect the normalization of the 3D mesh. If an isolated vertex is very far away from the 3D mesh model, the center of the bounding box is not the center of the 3D mesh after normalization, and the mesh is far away from the frame center (\emph{i.e.}, the center of the bounding box).
(c) Degenerate faces consist of three same vertices or three vertices are collinear. Degenerate faces and isolated vertices cannot provide any useful information for describing a shape and can be regarded as noise. What's more, they waste computing and storage resources.
(d) Self-intersecting faces are not a realistic representation of geological structures \cite{CaumonG}, and the intersections of these faces are not edges of the mesh.
(e) Inner surfaces are unnecessary because only the outer contour of the 3D mesh is required in general 3D computer vision applications. Besides, although the inner surfaces generally do not affect the outer contour of the 3D mesh, they affect the features that expected extracted only from the outer contour. The above geometric deficiencies overwhelm the advantages of meshes and become obstacles for utilizing meshes.

There are some existing methods to deal with the mesh, such as mesh denoising \cite{cascaded_normal} \cite{Yadav_denoising}, mesh surface simplification \cite{garland1997surface} and inside-outside classification of the 3D mesh model (\emph{e.g.}, ray casting-based method \cite{raycast}, winding number \cite{jacobson2013robust}, and a signed distance field-based method \cite{sdf}). These methods only process one type of many deficiencies and cannot eliminate many deficiencies in an end-to-end manner, and the deficiencies that affect mesh topology have not been eliminated.

To address the above issues and better make use of the advantages of the mesh, we design the corresponding operation to eliminate each type of deficiency. These operations are organized into an end-to-end mesh processing scheme to eliminate the mentioned five deficiencies thoroughly. The main contributions can be summarized as follows:
\begin{itemize}
\item
    An end-to-end scheme is proposed that can eliminate key deficiencies, \emph{i.e.}, remove duplicate elements, remove isolated vertices, remesh self-intersecting faces, and remove inner faces. So that we can utilize the advantages of mesh and save storage and computing resources.
\item
    Combining ray-casting and voting, we propose a new method that can effectively remove the inner faces of the 3D mesh. Simulating multiple random rays that shoot from a face, and accumulating the number of rays that can reach infinity. If the number is lower than a threshold, we can judge the face as an inner face and delete it.
\end{itemize}

\section{The Proposed Method}
A 3D mesh can be regarded as a graph that consists of vertices, edges and triangular faces, defined as $\mathcal{M}=\{\mathcal{V}, \mathcal{T}\}$. $\mathcal{V}=\{{v}_{i}\}^{N_v}_{i}$ represents the aggregation of $N_v$ vertices; $\mathcal{T}=\{t_{i}\}^{N_t}_{i=1}$ represents the aggregation of $N_t$ triangular faces, each face represents the 3-tuple of vertex indices $t_{i}=(p_{1}, p_{2}, p_{3})$, $1\leq{p_{1}, p_{2}, p_{3}}\leq{N_v}$ and consists of three edges $e^1_i=v_{p_2} - v_{p_1}$, $e^2_i=v_{p_3} - v_{p_2}$, and $e^3_i=v_{p_3} - v_{p_2}$.
In this section, we introduce these operations and the overall scheme in detail.

\textbf{Remove duplicate elements.} Duplicate elements are duplicate vertices and duplicate faces. A 3D mesh $\mathcal{M}$ may contain some vertices that satisfy:
\begin{equation}\label{eq:dup_v}
v_i=v_j,\space i\not= j, v_i, v_j \in \mathcal{V},
\end{equation}
then $v_i$ and $V_{\rm dup}=\{v_j\}$ are duplicate vertices. The duplicate faces are different faces $t_{i}=(p^i_{1}, p^i_{2}, p^i_{3})$ and $t_{j}=(p^j_{1}, p^j_{2}, p^j_{3})$ which have the same 3-tuple of vertices indices, but the order may be different, \emph{i.e.},
\begin{equation}\label{eq:dup_f}
\{p^i_{1}, p^i_{2}, p^i_{3}\}=\{p^j_{1}, p^j_{2}, p^j_{3}\}.
\end{equation}
After the duplicate elements are detected, we leave only one from the duplicate elements. The corresponding operation denoted as $f_{\rm dedup}$. The duplicate vertices should be removed before detecting and removing duplicate faces. Because after removing the duplicate vertices may produce new duplicate faces. Each 3-tuple of vertices indices for $\mathcal{T}$ need to be updated simultaneously when remove each duplicate vertex, because the vertices indices list changes when a vertex is removed.
%
\renewcommand{\algorithmicrequire}{\textbf{Input:}}
\renewcommand{\algorithmicensure}{\textbf{Output:}}

\begin{algorithm}[t]	
	\caption{\small Correct Face Orientation}
	\begin{algorithmic}[1]
		\REQUIRE
		$\mathcal{M}=\{\mathcal{V},  \mathcal{T}\}$, desired total rays $N_{\rm max}$, minimum rays $N_{\rm min}$ of each face;
		\ENSURE A repaired 3D mesh model $\mathcal{M}^{'}$;\\
		\STATE Compute area $s_i$ of per face $t_i$;
		\STATE $S = \sum_{i=1}^{T}s_i$;
		\FOR {$i = 1, ..., T$}
		\STATE $\textit{c}_{\rm front}^{i}=0$, $\textit{c}_{\rm back}^{i}=0$;
		\STATE $n_i ={\rm  max}(s_i/S*N_{\rm max},N_{\rm min})$;
		\STATE Randomly sample rays $\{r_{\rm front}^j, r_{\rm back}^j\}_{j=1}^{n_i}$ from the both sides of $t_i$. $r_{\rm front}^j$ and $r_{\rm back}^j$ have same origin and shoot in opposite directions;
		\FOR{$j = 1, ..., n_i$}
		\IF{$r_{\rm front}^j$ can reach infinity}
		\STATE $\textit{c}_{\rm front}^{i} = \textit{c}_{\rm front}^{i} + 1$;
		\ENDIF
		\IF{$r_{back}^j$ can reach infinity}
		\STATE $\textit{c}_{\rm back}^{i} = \textit{c}_{\rm back}^{i} + 1$;
		\ENDIF
		\ENDFOR
		\IF{$\textit{c}_{\rm front}^{i} <\textit{c}_{\rm back}^{i} $}
		\STATE flip $t_i$;
		\ELSE
		\STATE pass;
		\ENDIF
		\ENDFOR\\
		return $\mathcal{M}^{'}$;
	\end{algorithmic}
	\label{correct_facet}
\end{algorithm}

\textbf{Remove degenerate faces.} Degenerate faces are that some faces degenerate into a point or a line segment. They can be detected according to the following formula:
%
%
%
\begin{equation}\label{eq:degenerate2}
e^j_i\times e^k_i=0, \exists j,k\in t_i,  j\not=k.
\end{equation}
\emph{i.e.}, as for a degenerate face, the outer product of any two edges is zero. After detecting this deficiency, we remove the  3-tuple $t_i$ and keep the vertices of the  degenerate face. Because these vertices may be used by other faces, and they can be removed as isolated vertices if they are not used by other faces. This operation called $\textit{f}_{\rm remove\_deg}$.

\textbf{Remove isolated vertices.}
 Isolated vertex do not appear in any face's vertices, \emph{i.e.}, vertex $v_i$ satisfy :
\begin{equation}\label{eq:isolated}
 i \notin t_{i}=(p_{1}, p_{2}, p_{3}), \forall t_{i} \in \mathcal{T}.
\end{equation}
 These vertices can be deleted directly. We denote this operation as $\textit{f}_{\rm remove\_iso}$.
Similar to remove duplicate vertices, when removing each isolated vertex, each 3-tuple of vertices indices for $\mathcal{T}$ need to be updated simultaneously.
\begin{figure*}
	\centering
	
	\subfigure[Original mesh model.]
	{
		\label{fig:airplane_635_Original} 
		\includegraphics[width=1.6in]{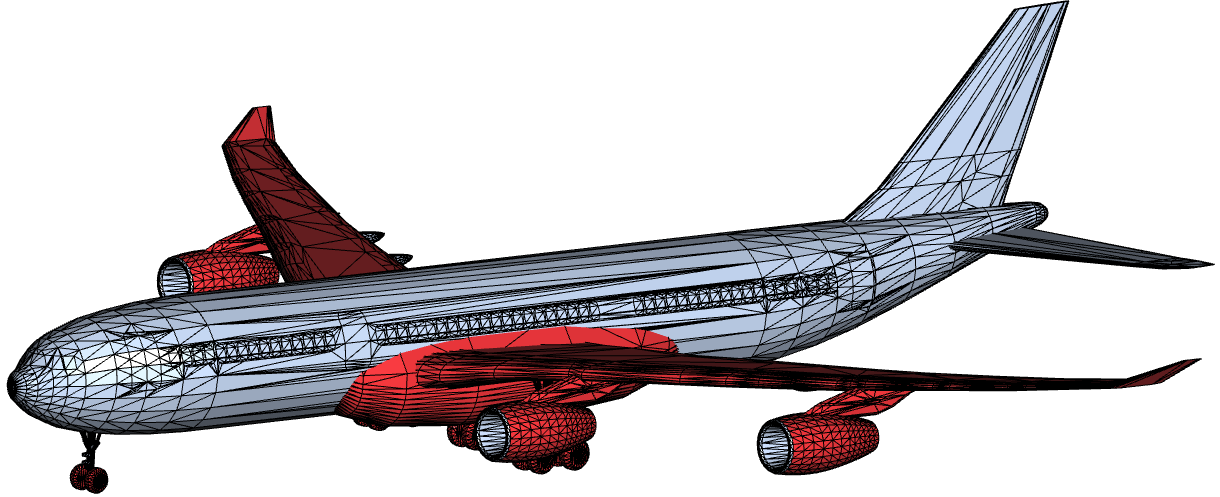}
	}
	\subfigure[Deduplicate mesh model.]
	{
		\label{fig:airplane_635_dup} 
		\includegraphics[width=1.6in]{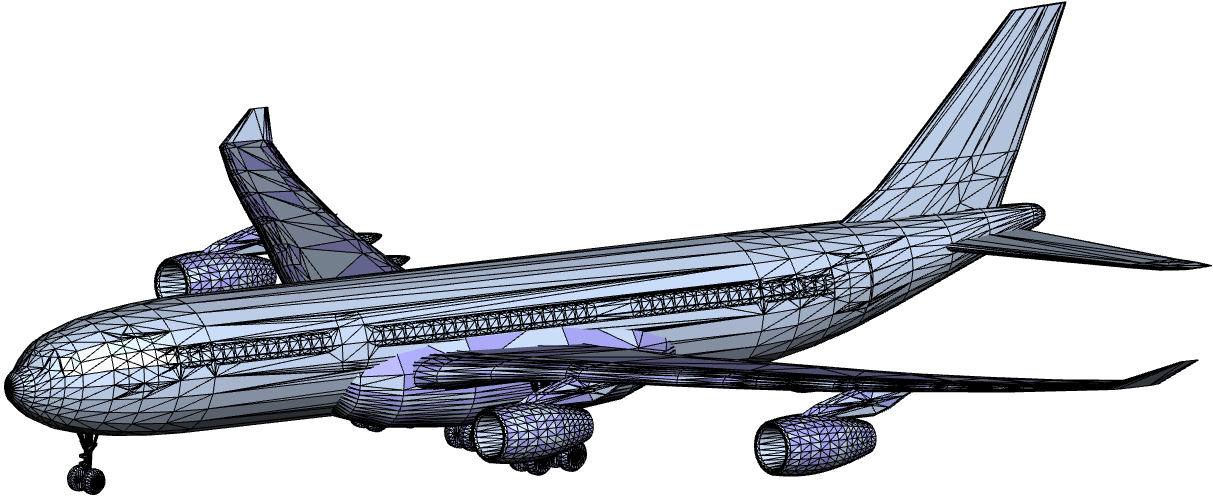}
	}
	\subfigure[Self-intersecting faces.]
	{
		\label{fig:airplane_635_dup_SI} 
		\includegraphics[width=1.6in]{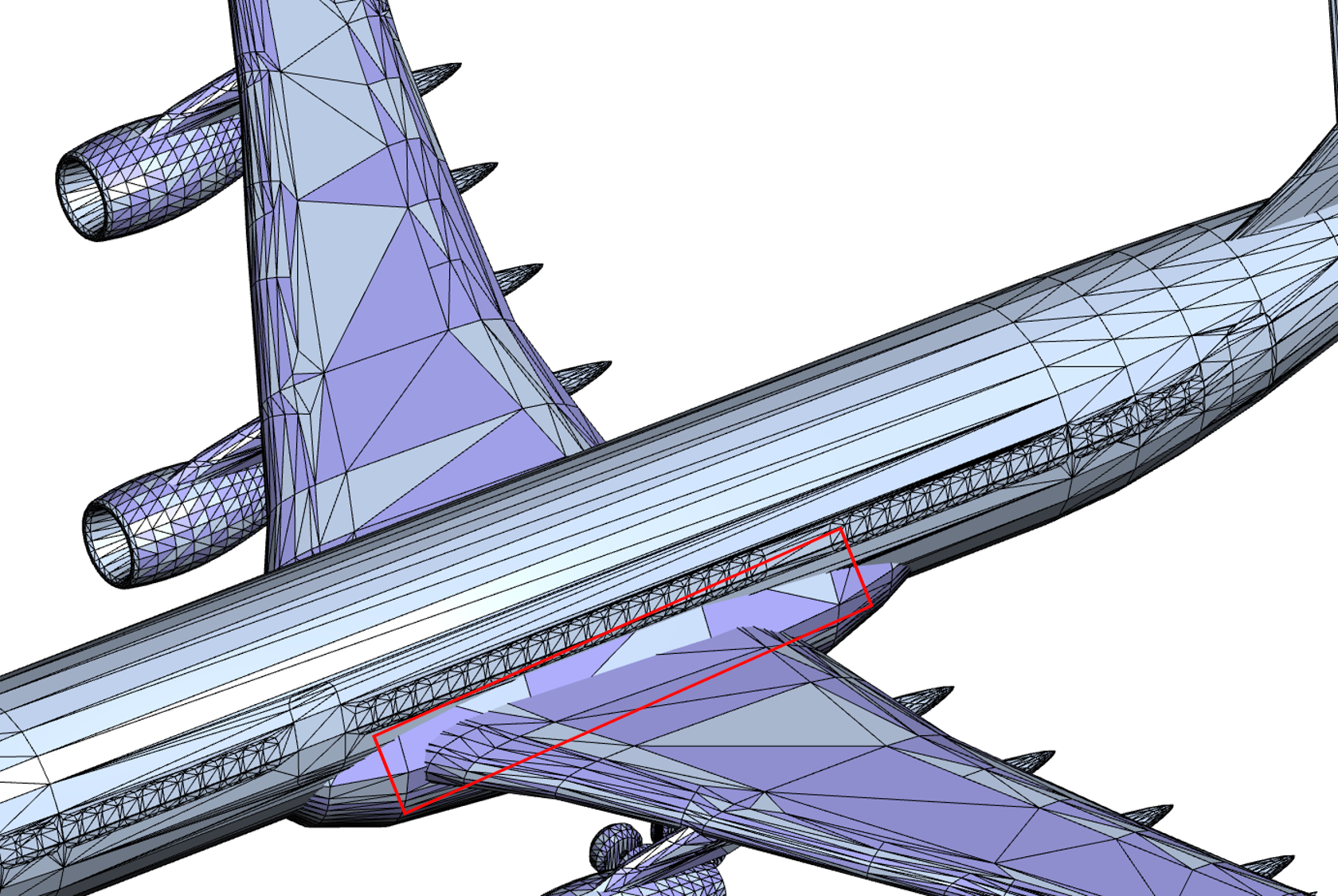}
	}
	\subfigure[Remeshed self-intersecting faces.]
	{
		\label{fig:airplane_635_dup_remeshSIbb} 
		\includegraphics[width=1.6in]{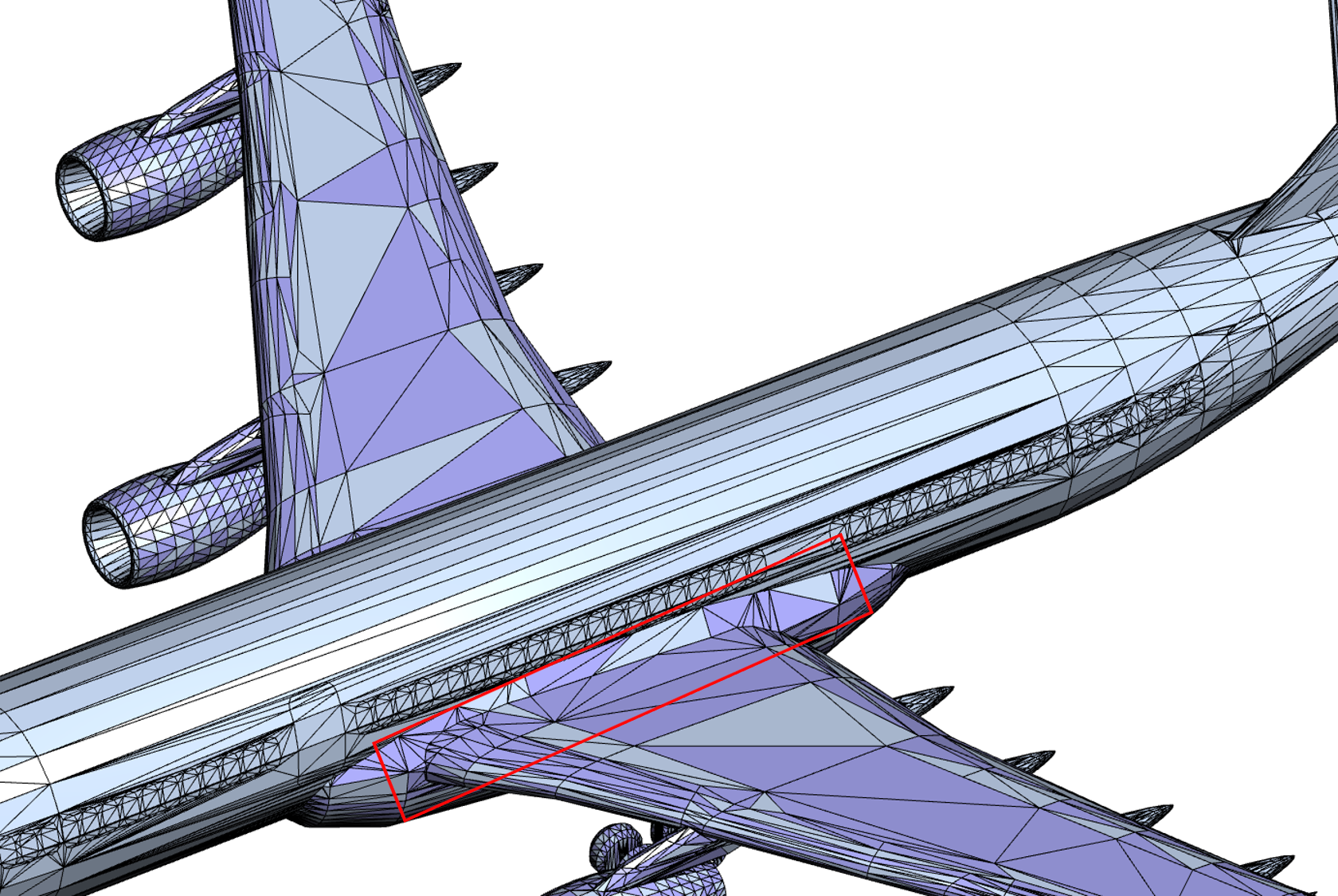}
	}
	\subfigure[Simplified mesh model.]
	{
		\label{fig:airplane_635_dup_remeshSI_simplify10000_mesh} 
		\includegraphics[width=1.6in]{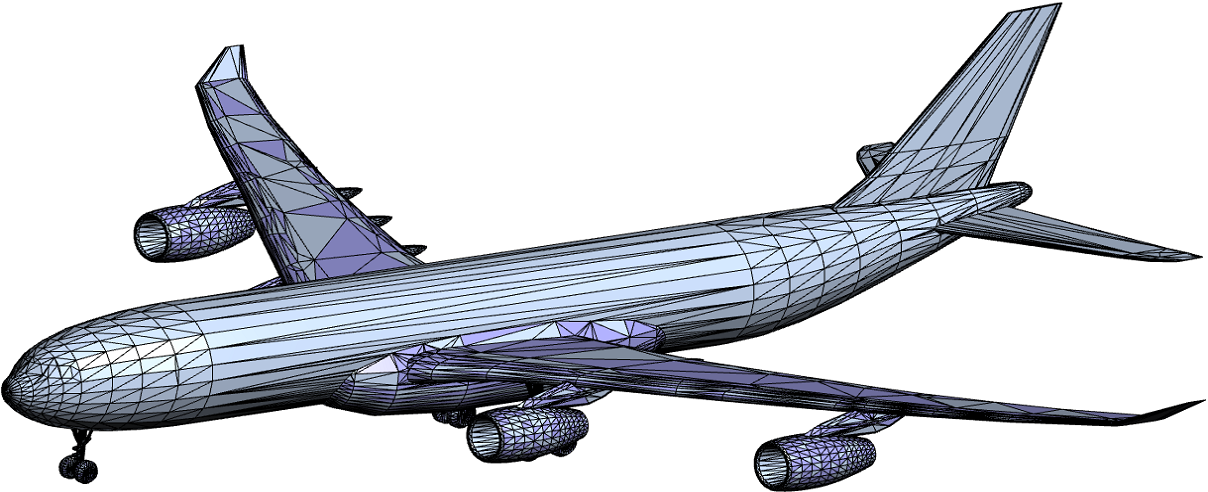}
	}
	\subfigure[Correct faces' orientation.]
	{
		\label{fig:airplane_635_dup_remeshSI_simplify10000_correctface_mesh} 
		\includegraphics[width=1.6in]{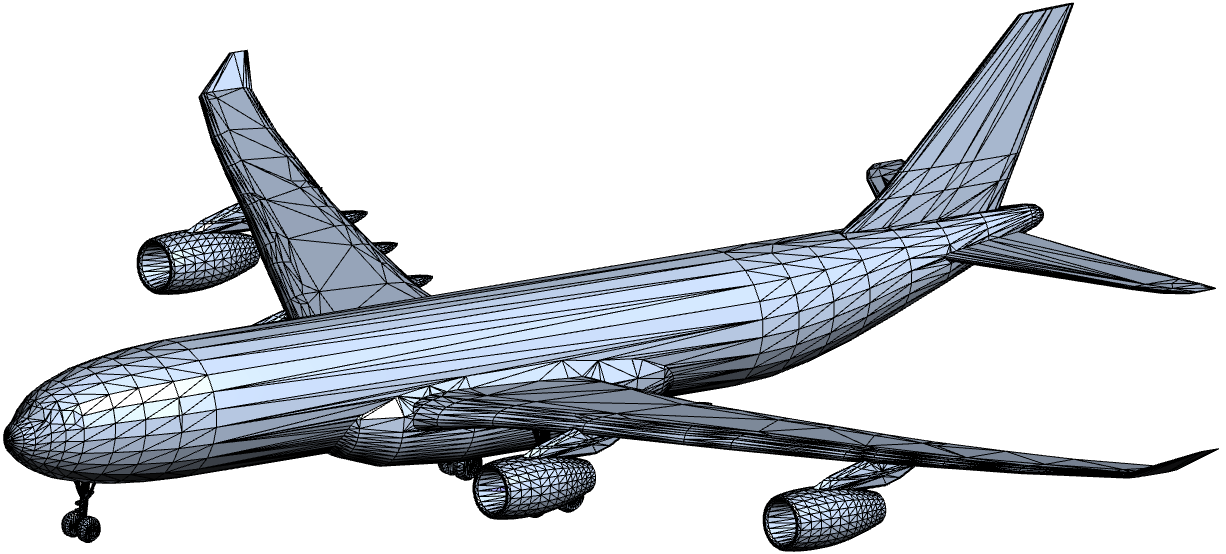}
	}
	\subfigure[Inner faces.]
	{
		\label{fig:airplane_635_inner} 
		\includegraphics[width=1.6in]{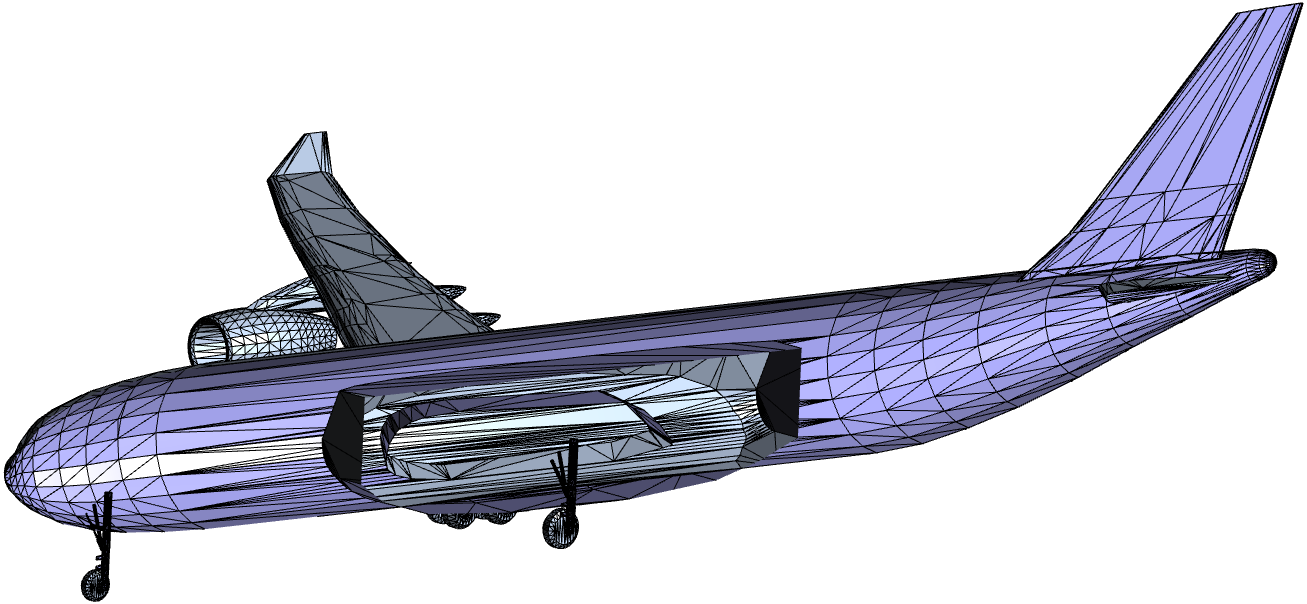}
	}
	\subfigure[Slice of after removing inner faces.]
	{
		\label{fig:airplane_635_remove_inner} 
		\includegraphics[width=1.6in]{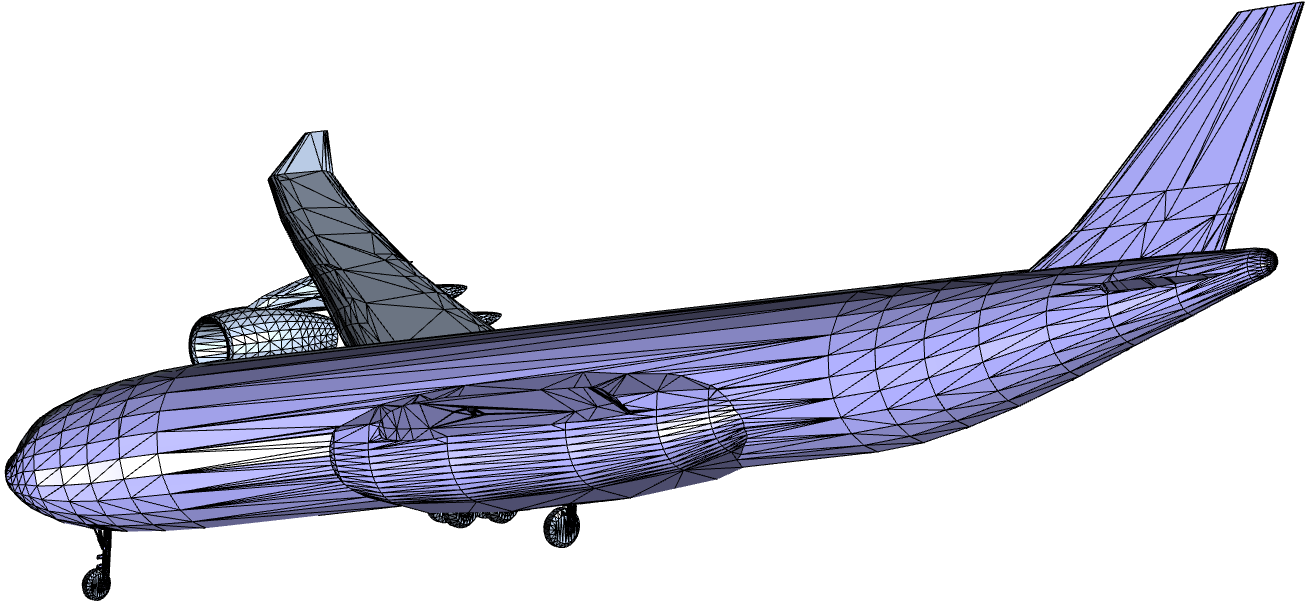}
	}
	\caption{Visualization of the intermediate processing results for airplane\_0635 in ModelNet40 \cite{modelnet}. Gray face indicates outside	and light blue face indicates inside. Processing sequence: (a) Original mesh model ( red part is duplicate elements ). (b) Mesh model is processed via $\textit{f}_{\rm dedup}$, $\textit{f}_{\rm remove\_deg}$ and $\textit{f}_{\rm remove\_iso}$. Light blue face means its orientation is incorrect. (c) A local patch of self-intersecting faces. (d) Self-intersecting faces are repaired by remeshing self-intersections algorithm. (e) We simplify the mesh model to speed up the processing of subsequent steps. (f) Mesh model of previous steps exists wrong face orientation, which is corrected via \textbf{Algorithm \ref{correct_facet}}. (g) A cross-section view of the model to reveal its interior structure. (h) A cross-section view of the model after removing inner faces, which proves that we can remove the inner faces thoroughly.}
	\label{fig:airplane_635} 
\end{figure*}

\textbf{Remesh self-intersections.} The self-intersecting faces can be repaired by remeshing them. We use the remeshing algorithm implemented Mapple \cite{mapple}, which uses an AABB tree, to achieve this.
This algorithm detects the intersection of two faces firstly. If it is a common edge of both faces, the algorithm detects the next pair faces. If it is not a common edge, taking an appropriate number of points at equal intervals on it and triangulating two faces separately. We denote this operation as $\textit{f}_{\rm remesh\_si}$.

\renewcommand{\algorithmicrequire}{\textbf{Input:}}
\renewcommand{\algorithmicensure}{\textbf{Output:}}
\begin{algorithm}[t]	
	\caption{\small Remove The Inner Faces}
	\begin{algorithmic}[1]
		\REQUIRE
		$\mathcal{M}=\{\mathcal{V},  \mathcal{T}\}$, desired total rays $N_{\rm max}$, minimum rays $N_{\rm min}$ of each face;
		\ENSURE A repaired and 3D mesh model $\mathcal{M}^{'}$;\\
		\STATE Compute area $s_i$ of per face $t_i$;
		\STATE $S = \sum_{i=1}^{T}s_i$;
		\FOR {$i = 1, ..., T$}
		\STATE $\textit{c}_{\rm inf}^{i}=0$;
		\STATE $n_i ={\rm max}(s_i/S*N_{\rm max}, N_{\rm min})$;
		\STATE Randomly sample rays $\{r_j\}_{j=1}^{n_i}$ from the outside of $t_i$;	
		\FOR{$j = 1, ..., n_i$}
		\IF{$r_j$ can reach infinity}
		\STATE $\textit{c}_{\rm inf}^{i} = \textit{c}_{\rm inf}^{i} + 1$;
		\ENDIF
		\ENDFOR
		\IF{$\textit{c}_{\rm inf}^i< 0.05\textit{n}_{i}$}
		\STATE delete $t_i$;
		\ENDIF
		\ENDFOR\\
		\STATE $\mathcal{M^{'}}=\textit{f}_{\rm remove\_iso}(\mathcal{M})$;\\
		return $\mathcal{M}^{'}$;
	\end{algorithmic}
	\label{remove_inner}
\end{algorithm}

\textbf{Remove inner faces.} The orientation of some faces are incorrect after the operation $\textit{f}_{\rm dedup}$ or are incorrect original, for instance, the mesh model of Fig. \ref{fig:si}, where gray indicates outside and light blue indicates inside. Before removing the inner faces of the 3D mesh, correcting facets orientation is necessary and can make visualization better. The method of \cite{takayama2014simple} can solve it well, which is summarized in \textbf{Algorithm \ref{correct_facet}}. Specifically, for each face $t_i$, randomly sampling a large number of rays, whose origin and direction are both random and number proportional to the area of the face. Besides, two rays are shot in opposite directions for each ray origin and direction, \emph{i.e.}, $r^j_{\rm front}$ and $r^j_{\rm back}$. The corresponding counter $\textit{c}_{\rm front}^{i}$ (resp. $\textit{c}_{\rm back}^{i}$) is incremented if a ray shot the from front (resp. back) side of the face does not intersect with any other
faces. After shooting all the rays, the face $t_i$ is flipped if $\textit{c}_{\rm front}^{i} < \textit{c}_{\rm back}^{i}$, \emph{i.e.},  $t_{i}=(p_{1}, p_{2}, p_{3}) \rightarrow t_{i}=(p_{1}, p_{3}, p_{2})$.

\renewcommand{\algorithmicrequire}{\textbf{Input:}}
\renewcommand{\algorithmicensure}{\textbf{Output:}}
\begin{algorithm}[t]
	\caption{\small Eliminating Geometric Deficiencies of a 3D Mesh}
	\begin{algorithmic}[1]
		\REQUIRE 
		$\mathcal{M}=\{\mathcal{V},  \mathcal{T}\}$, expected number $N_{d};
		$
		\ENSURE A repaired and simplified 3D mesh $\mathcal{M}^{'}$;\\
		\STATE $\mathcal{M}=\textit{f}_{\rm dedup}(\mathcal{M})$;
		\STATE $\mathcal{M}=\textit{f}_{\rm remove\_deg}(\mathcal{M})$;
		\STATE $\mathcal{M}=\textit{f}_{\rm remove\_iso}(\mathcal{M})$;
		\STATE Normalize $\mathcal{M}$ into a unit sphere;
		\STATE $\mathcal{M}=\textit{f}_{\rm remesh\_si}(\mathcal{M})$;
		\STATE $\mathcal{M}=\textit{f}_{\rm dedup}(\mathcal{M})$;
		\STATE $\mathcal{M}=\textit{f}_{\rm simplify}(\mathcal{M}, N_{d})$;
		\STATE Correct each facet's orientation \textbf{Algorithm \ref{correct_facet}}, get $\mathcal{M}$;
		\STATE Remove the inner structure via \textbf{Algorithm \ref{remove_inner}}, get $\mathcal{M}^{'}$;
		return $\mathcal{M}^{'}$;
	\end{algorithmic}
	\label{Pipeline}
\end{algorithm}

\begin{figure*}
	\centering	
	\subfigure[Original mesh model.]
	{
		\label{fig:guitar_0205_ori} 
		\includegraphics[width=1.5in]{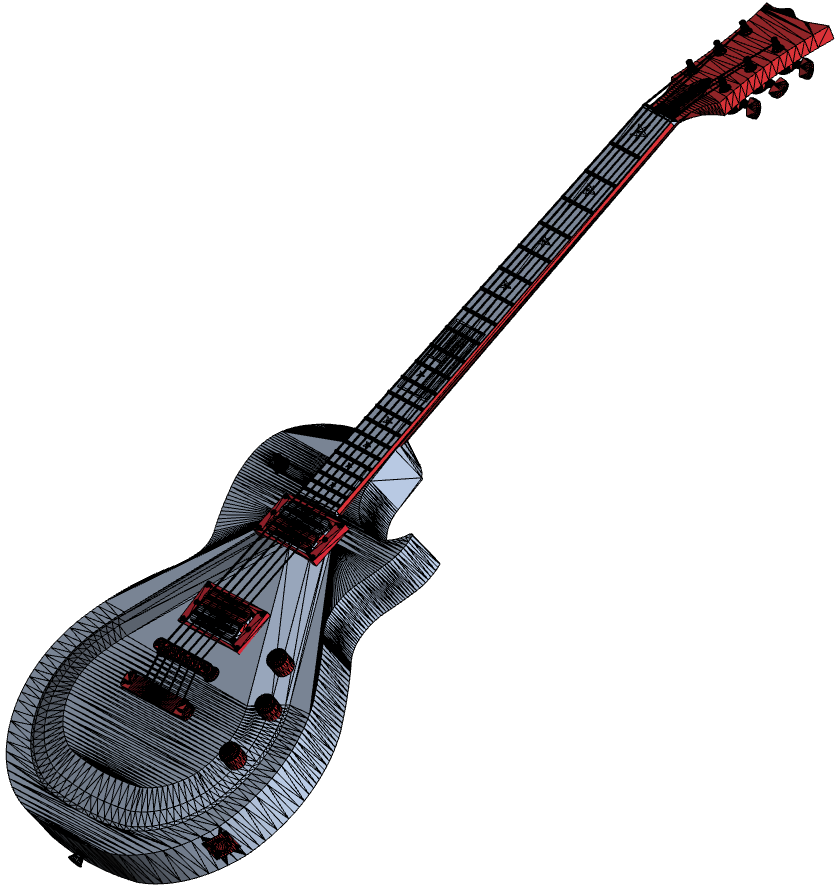}
	}
	\subfigure[Deduplicate mesh model.]
	{
		\label{fig:guitar_0205_dup} 
		\includegraphics[width=1.5in]{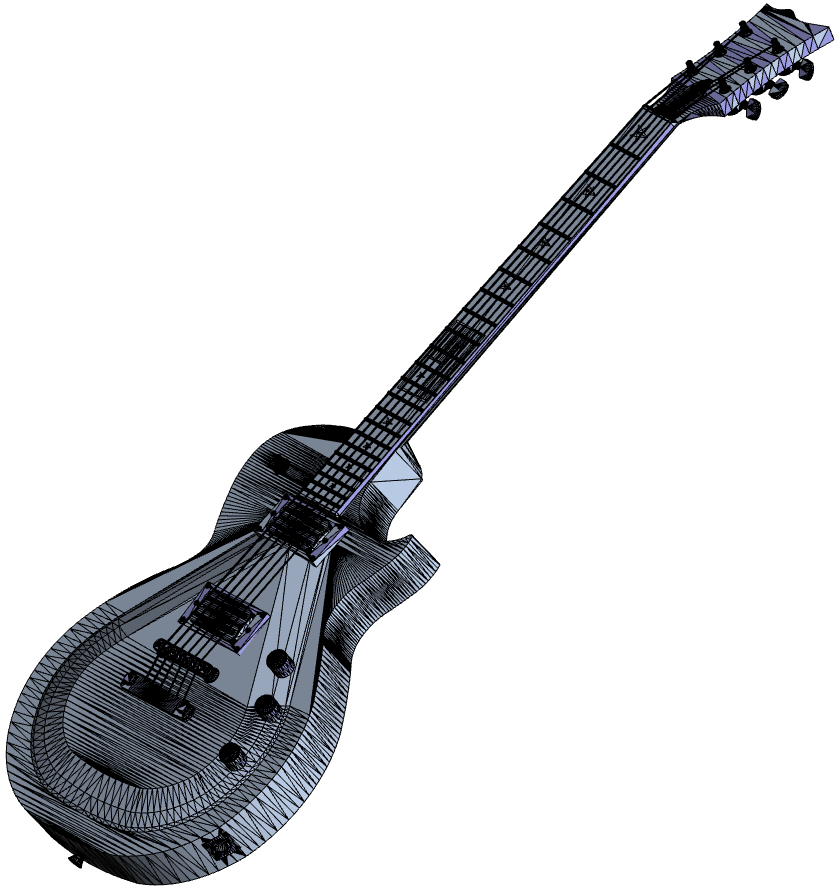}
	}
	\subfigure[Self-intersecting faces.]
	{
		\label{fig:guitar_0205_dup_SI} 
		\includegraphics[width=1.5in]{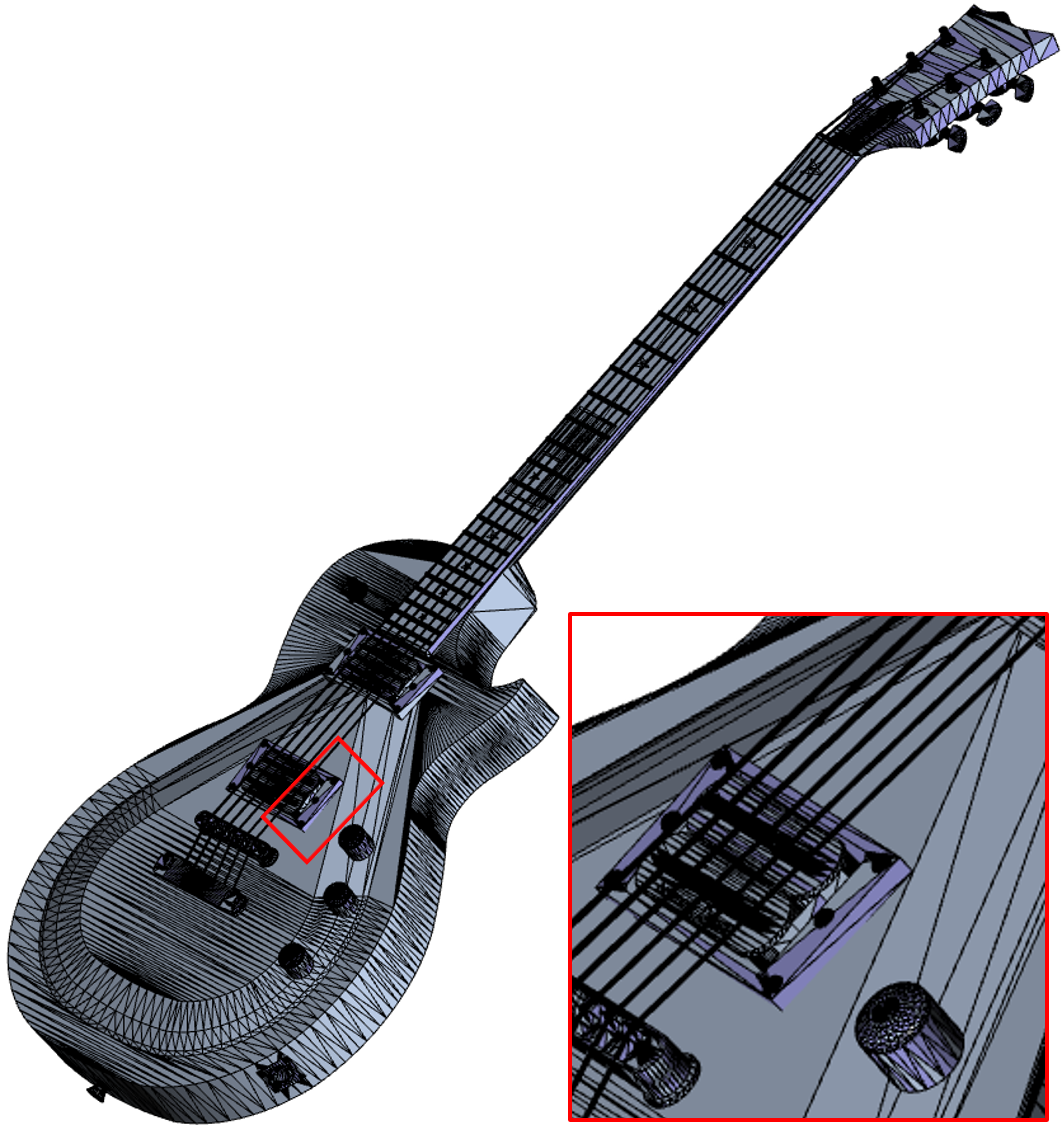}
	}
	\subfigure[Remeshed self-intersecting faces.]
	{
		\label{fig:guitar_0205_dup_remeshSI} 
		\includegraphics[width=1.6in]{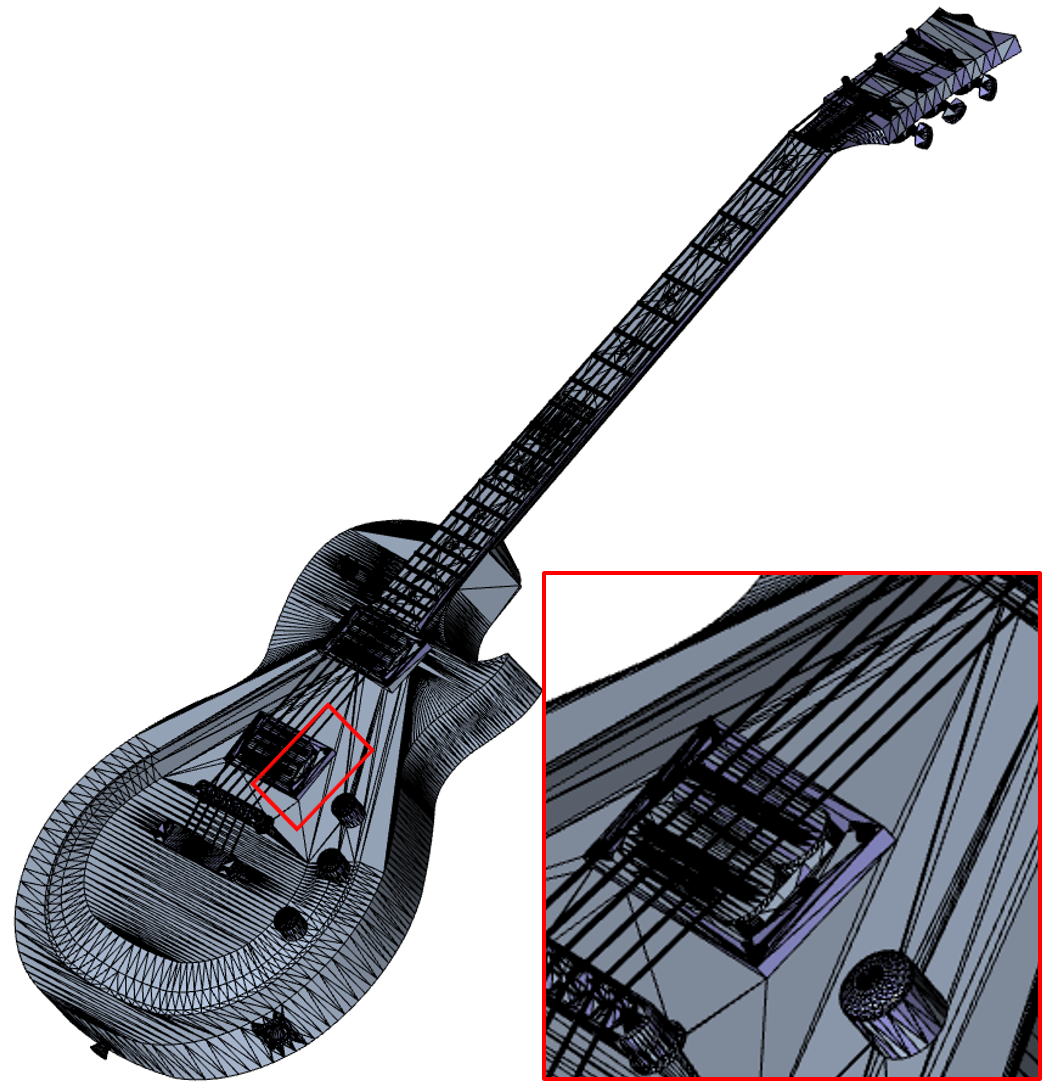}
	}
	\subfigure[Simplified mesh model.]
	{
		\label{fig:guitar_0205_dup_remeshSI_simplify} 
		\includegraphics[width=1.5in]{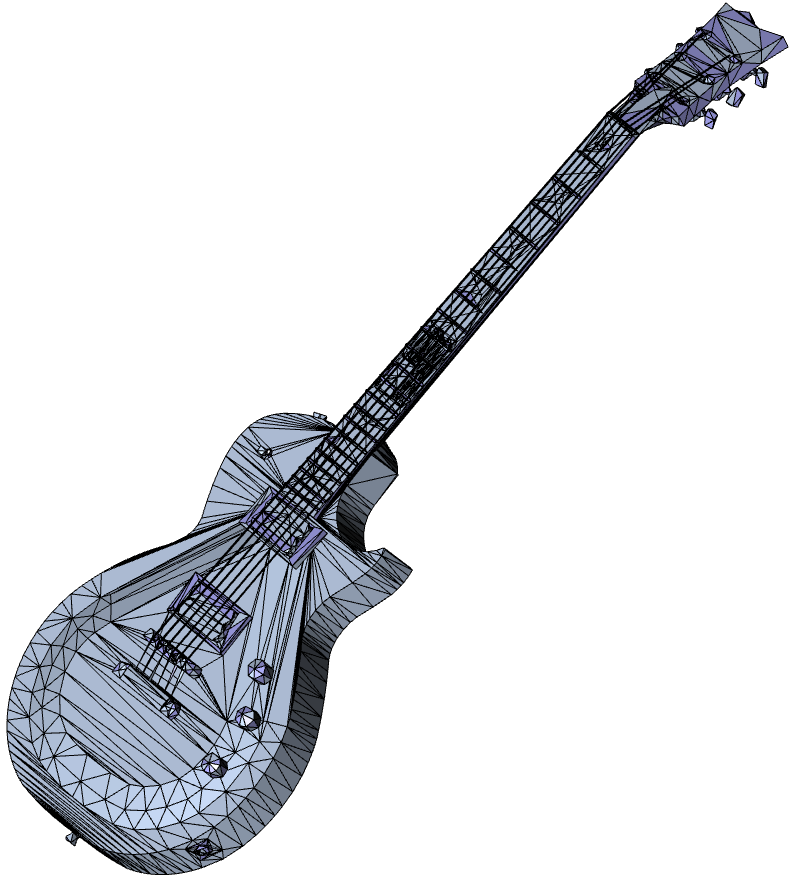}
	}
	\subfigure[Correct faces orientation.]
	{
		\label{fig:guitar_0205_dup_remeshSI_simplify_remove_inner} 
		\includegraphics[width=1.5in]{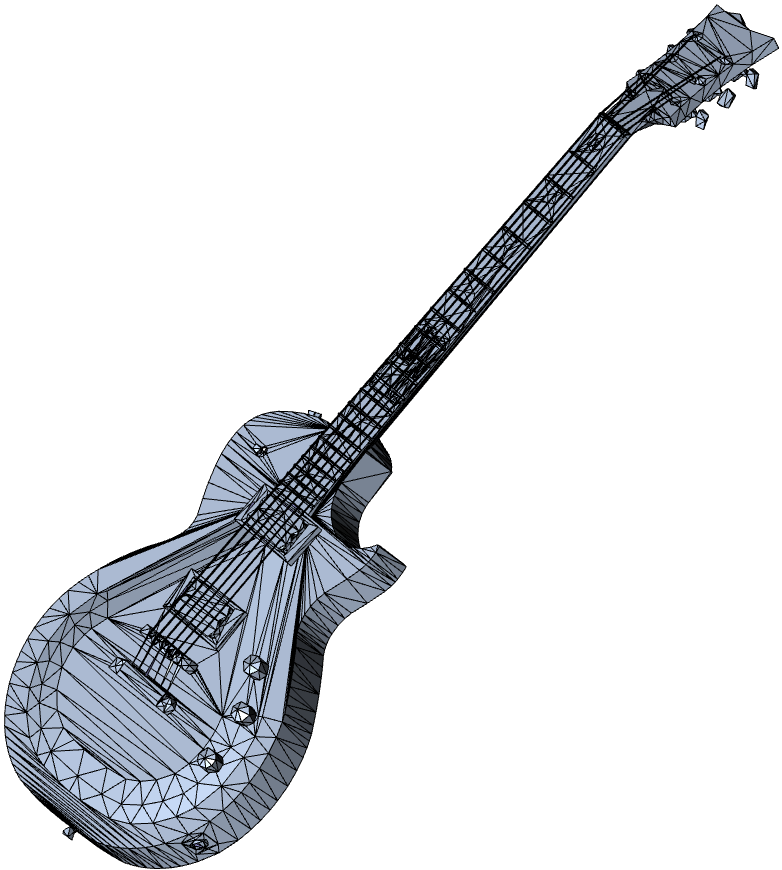}
	}
	\subfigure[Inner faces.]
	{
		\label{fig:guitar_inner} 
		\includegraphics[width=1.5in]{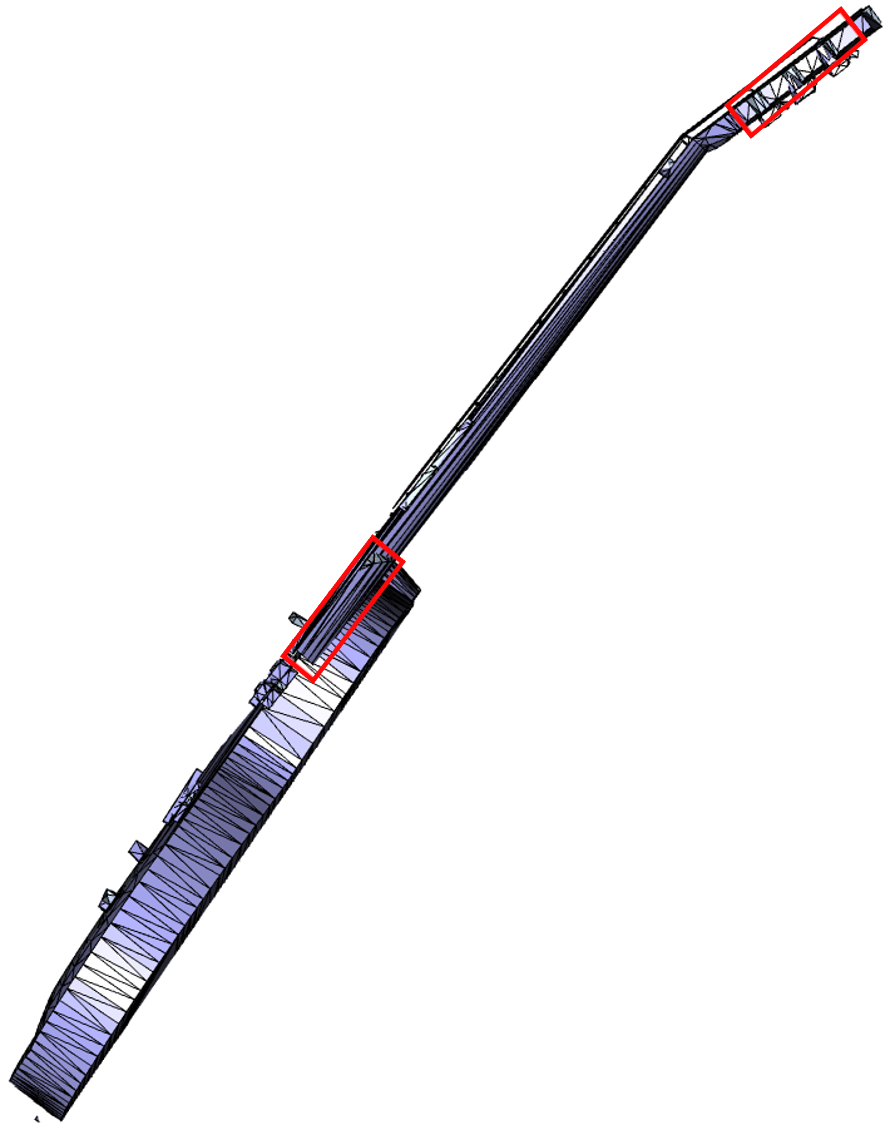}
	}
	\subfigure[Slice of after removing inner faces.]
	{
		\label{fig:guitar_remove_inner} 
		\includegraphics[width=1.5in]{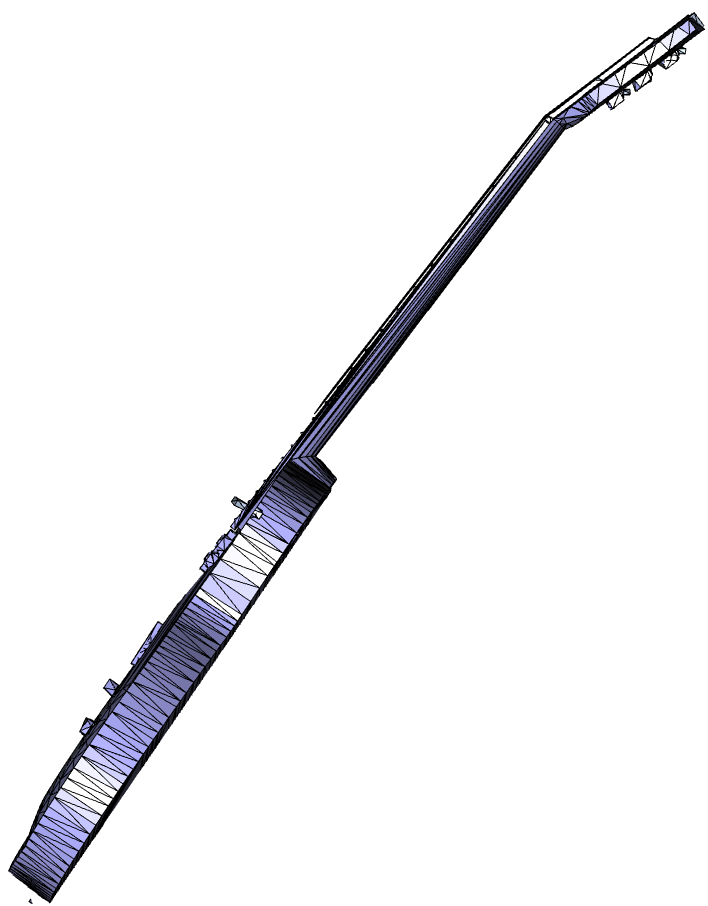}
	}
	\caption{Visualization of the intermediate processing results for guitar\_0205 in ModelNet40 \cite{modelnet}. Gray face indicates outside and light blue face indicates inside. Processing sequence: (a) Original mesh model ( red part is duplicate elements ). (b) Mesh model processed via $\textit{f}_{dedup}$, $\textit{f}_{remove\_deg}$ and $\textit{f}_{remove\_iso}$. Light blue face means its orientation is incorrect. (c) A local patch of self-intersecting faces. (d) Self-intersecting faces are repaired by remeshing self-intersections algorithm. (e) We simplify the mesh model to speed up the processing of subsequent steps. (f) Mesh model of previous steps exists wrong face orientation, which is corrected via \textbf{Algorithm \ref{correct_facet}}. (g) A cross-section view of the model to reveal its interior structure. (h) A cross-section view of the model after removing inner faces, which proves that we can remove the inner faces thoroughly.}
	\vspace{-10pt}
	\label{fig:guitar_0205} 
\end{figure*}

For detecting and removing inner faces of the 3D mesh, a method based on ray casting and voting that inspired by~\cite{takayama2014simple} is proposed, which is summarized in \textbf{Algorithm \ref{remove_inner}}. The main idea is to judge each face whether visible from the outside or not, which is decided by randomly shooting many rays from the outside of the face. Specifically, for every face $\textit{t}_{i}$, we randomly sample $\textit{n}_{i}$ ($\textit{n}_{i}\ge100$) points proportional to the face's area from the outside of the face as the ray origins, and the direction of each ray is also randomly sampled. If a ray does not intersect with any other faces in the 3D mesh, \emph{i.e.}, it can reach infinity,  which indicates the corresponding face is an outer face, the corresponding score counter $\textit{c}_{\rm inf}^i$ is incremented. After shooting all rays, the face can be regarded as an inner face if $\textit{c}_{\rm inf}^i< 0.05\textit{n}_{i}$. Then 3-tuple $t_i$ can be removed, while keeping its vertices that may be used by outer faces. After removing all inner faces, $\textit{f}_{\rm remove\_iso}$ is conducted again to remove those vertices that only used by inner faces.

\textbf{Overall Scheme.}
 The overall processing scheme is shown in \textbf{Algorithm~\ref{Pipeline}}. First,  operations $\textit{f}_{\rm dedup}$, $\textit{f}_{\rm remove\_deg}$ and $\textit{f}_{\rm remove\_iso}$ are conducted one by one to remove duplicate elements, degenerate faces, and isolated vertices successively. Second, we normalize the mesh model into a unit sphere for the convenience of visualization in subsequent steps. Afterward,  $\textit{f}_{\rm remesh\_si}$ is conducted to remesh self-intersecting faces. $\textit{f}_{\rm remesh\_si}$ may produce duplicate elements, so operation $\textit{f}_{\rm dedup}$ is conducted again.
 Some mesh models may have a large number of vertices, which costs much processing time. Thus the mesh is simplified to the expected number of vertices by mesh simplification algorithm \cite{garland1997surface}, which can speed up the processing of subsequent steps and is denoted as $\textit{f}_{\rm simplify}(\mathcal{M}, N_{d})$, where $N_{d}$ is the expected number. Finally, the face orientation is corrected via \textbf{Algorithm \ref{correct_facet}} and the undesired inner faces are removed via \textbf{Algorithm \ref{remove_inner}}.
\section{Experiment Results}
 We implement the overall processing scheme in C++ based on the Libigl library \cite{libigl}, and verify the validity of the proposed method on ModelNet40 \cite{modelnet}.
 The intermediate processing results of  airplane\_0635 and guitar\_0205 via \textbf{Algorithm \ref{Pipeline}} are shown in  Fig. \ref{fig:airplane_635} and Fig. \ref{fig:guitar_0205} respectively.

 In this section, we introduce the intermediate processing results of airplane\_0635 in detail.
 The original mesh model has 25,085 vertices and 31,222 faces and is illustrated in Fig. \ref{fig:airplane_635_Original}, and the red part is duplicate elements. Fig. \ref{fig:airplane_635_dup} shows the mesh model processed via~$\textit{f}_{\rm dedup}$, ~$\textit{f}_{\rm remove\_deg}$ and~$\textit{f}_{\rm remove\_iso}$ (\emph{i.e.}, step 1-3 of \textbf{Algorithm \ref{Pipeline}}). Its number of vertices and faces decrease to 11,492 and 22,528 respectively. Comparing with the original mesh model, the number of vertices reduces more than half. The results illustrate that the original mesh may contain massive duplicate elements, degenerate faces and isolated vertices that waste much storage and computing resources when saving it to disk or conduct some computing tasks on it. Then,  self-intersecting faces are remeshed. Comparing Fig. \ref{fig:airplane_635_dup_SI}  and Fig. \ref{fig:airplane_635_dup_remeshSIbb}, the self-intersecting faces can be remeshed well, which provides a guarantee for effective removal of the inner faces. Then we conduct~$\textit{f}_{\rm dedup}$ and~$\textit{f}_{\rm simplify}$,
  and the result is shown in Fig. \ref{fig:airplane_635_dup_remeshSI_simplify10000_mesh}. Contrasting Fig. \ref{fig:airplane_635_dup_remeshSI_simplify10000_mesh} and Fig. \ref{fig:airplane_635_dup_remeshSIbb}, we can find some small and unimportant faces (\emph{e.g.}, the windows of the airplane) are simplified/disappeared, which can speed up subsequent processing steps. In addition, the orientation of some faces is wrong in Fig. \ref{fig:airplane_635_dup} - Fig. \ref{fig:airplane_635_dup_remeshSI_simplify10000_mesh} (\emph{i.e.}, gray indicates outside and light blue indicates inside). So \textbf{Algorithm \ref{correct_facet}} is used to correct the orientation of these faces, and the result is shown in Fig. \ref{fig:airplane_635_dup_remeshSI_simplify10000_correctface_mesh}. Then we remove the inner structure via \textbf{Algorithm \ref{remove_inner}}.
  The slices before and after removing the inner faces are shown in Fig.  \ref{fig:airplane_635_inner} and \ref{fig:airplane_635_remove_inner}, respectively,
  which proves the proposed \textbf{Algorithm \ref{remove_inner}} can effectively remove the inner faces. Specially, the mesh owns 10,000 vertices and 23,252 faces before removing inner faces (\emph{i.e.}, Fig. \ref{fig:airplane_635_inner}), and the mesh owns 8,564 and 16,189 faces after removing inner faces (\emph{i.e.}, Fig. \ref{fig:airplane_635_remove_inner}), which indicates remove inner faces can save resources for other tasks that consume mesh. In the future, the 3D mesh processed by our method could be efficiently apllied into many computer vision fieleds. \emph{e.g.}, object recognition~\cite{cong2018speedup,liu2020robust}, domain adaptation~\cite{ Dong_2019_ICCV}.

\section{Conclusion}
In this paper, we propose an end-to-end processing scheme to effectively eliminate the deficiencies of 3D meshes, then we can utilize the advantages of deficiency-free meshes in other tasks (\emph{e.g.}, object reconstruction, scene reconstruction, and object recognition). In addition, a new method is proposed that combines ray-casting and voting and removes the inner faces of 3D mesh effectively. The intermediate processing results demonstrate the effectiveness of our method. In the future, we plan to research 3D mesh object recognition and 3D mesh scene understanding, for which an effective preprocessing scheme and high quality 3D meshes are necessary for these tasks.


\begin{thebibliography}{00}
\bibitem{pixel2mesh}
	Wang, Nanyang, et al, ``Pixel2mesh: Generating 3d mesh models from single rgb images,'' Proceedings of the European Conference on Computer Vision (ECCV), 2018, pp. 52-67.
\bibitem{3drcnn}
	Kundu, Abhijit, Yin Li, and James M. Rehg, ``3d-rcnn: Instance-level 3d object reconstruction via render-and-compare." Proceedings of the IEEE Conference on Computer Vision and Pattern Recognition. 2018, pp.3559-3568.
\bibitem{meshPiazza2018}
	Piazza, Enrico, Andrea Romanoni, and Matteo Matteucci, ``Real-time cpu-based large-scale three-dimensional mesh reconstruction," IEEE Robotics and Automation Letters, vol.3, no.3, pp. 1584-1591, 2018
\bibitem{Rosinol_meshre}
	Rosinol, Antoni, et al, ``Incremental Visual-Inertial 3D Mesh Generation with Structural Regularities." arXiv preprint arXiv:1903.01067 (2019).
\bibitem{Deepshape}
	Xie, Jin, et al, ``Deepshape: Deep-learned shape descriptor for 3D shape retrieval," IEEE transactions on pattern analysis and machine intelligence, vol. 39, no. 7, 2016. 1335-1345.
\bibitem{wavekernel}
	Aubry, Mathieu, Ulrich Schlickewei, and Daniel Cremers, ``The wave kernel signature: A quantum mechanical approach to shape analysis," IEEE international conference on computer vision workshops (ICCV workshops), 2011, pp. 1626-1633.
\bibitem{L3DOC}
    Liu, Yuyang, Yang Cong, and Gan Sun, ``L3DOC: Lifelong 3D Object Classification," arXiv preprint arXiv:1912.06135(2019).
\bibitem{modelnet}	
	Wu, Zhirong, et al, ``3d shapenets: A deep representation for volumetric shapes," Proceedings of the IEEE conference on computer vision and pattern recognition, 2015, pp. 1912-1920.
\bibitem{CaumonG}
	Caumon G, Collon-Drouaillet P, De Veslud C L C, et al, ``Surface-based 3D modeling of geological structures," Mathematical Geosciences, vol.41, no.8, pp.927-94, 2009.
\bibitem{shapenet}
	Chang, Angel X., et al, ``Shapenet: An information-rich 3d model repository," arXiv preprint arXiv:1512.03012 (2015).
\bibitem{takayama2014simple}
	Takayama, Kenshi, et al, ``A simple method for correcting facet orientations in polygon meshes based on ray casting," Journal of Computer Graphics Techniques, vol. 3, no. 3, pp.53-63, 2014.
\bibitem{cascaded_normal}
Wang, Peng-Shuai, Yang Liu, and Xin Tong, ``Mesh denoising via cascaded normal regression," ACM Transactions on Graphics (TOG), vol. 35, no. 6,  pp. 1-12, 2016.
\bibitem{Yadav_denoising}
Yadav, Sunil Kumar, Ulrich Reitebuch, and Konrad Polthier. ``Robust and high fidelity mesh denoising." IEEE transactions on visualization and computer graphics vol. 25, no. 6, pp.2304-2310, 2018.
\bibitem{garland1997surface}	
	Garland, Michael, and Paul S. Heckbert, ``Surface simplification using quadric error metrics," Proceedings of the 24th annual conference on Computer graphics and interactive techniques, ACM Press/Addison-Wesley Publishing Co, 1997, pp. 209-216.
\bibitem{mapple}
	 L. Nan, ``Mapple," https://3d.bk.tudelft.nl/liangliang/software.html.
\bibitem{raycast}
	Nooruddin, Fakir S., and Greg Turk. ``Interior/exterior classification of polygonal models." Proceedings Visualization. VIS (Cat. No. 00CH37145). IEEE, 2000, pp.415-422.
\bibitem{jacobson2013robust}
	Jacobson, Alec, Ladislav Kavan, and Olga Sorkine-Hornung, ``Robust inside-outside segmentation using generalized winding numbers." ACM Transactions on Graphics (TOG), vol. 32, no.4, pp.33:1-33:12, 2013.
\bibitem{libigl}
	Alec Jacobson, et al, ``libigl," https://github.com/libigl/libigl.
\bibitem{sdf}
	Xu, Hongyi, and Jernej Barbi. ``Signed distance fields for polygon soup meshes." Proceedings of Graphics Interface. Canadian Information Processing Society, 2014, pp. 35-41.
\bibitem{Embree}
	``Embree," https://www.embree.org/.
\bibitem{cong2018speedup}
Cong, Yang, et al. "Speedup 3-D texture-less object recognition against self-occlusion for intelligent manufacturing." IEEE transactions on cybernetics, vol. 49, no. 11, pp. 3887–3897, 2018.
\bibitem{liu2020robust}
Liu, Hongsen, et al. "Robust 3-D Object Recognition via View-Specific Constraint." IEEE Transactions on Systems, Man, and Cybernetics: Systems, 2020.
%
\bibitem{Dong_2019_ICCV}	
	Dong, Jiahua, et al. "Semantic-transferable weakly-supervised endoscopic lesions segmentation." Proceedings of the IEEE International Conference on Computer Vision, 2019, pp. 10712-10721.

\end{thebibliography}
\end{document}